\pdfoutput=1
\documentclass[11pt]{article}
\usepackage{listings}

\usepackage{ACL2023}
\usepackage{array}

\usepackage{tcolorbox}
\usepackage{times}
\usepackage{latexsym}
\usepackage{import}
\usepackage{graphicx}
\usepackage{booktabs}
\usepackage{amssymb}
\usepackage{multirow}
\usepackage{blindtext}
\usepackage{enumitem}
\usepackage{xcolor}
\usepackage{xparse}
\DeclareDocumentCommand\dia{ o m }{%
    \begin{itemize}[%
        ,label=\IfNoValueTF {#1} {}{#1:}
        ,labelsep=8mm
        ,nosep % optional
        ,font=\color{black}
        ]
        \item #2
    \end{itemize}%  
    }

\definecolor{darkgreen}{rgb}{112,114,88}

\usepackage[T1]{fontenc}
\usepackage[utf8]{inputenc}

\usepackage{microtype}
\newcolumntype{P}[1]{>{\centering\arraybackslash}p{#1}} %% horizontal centering

\usepackage{inconsolata}

\usepackage{xcolor}

\definecolor{codegreen}{rgb}{0,0.6,0}
\definecolor{codegray}{rgb}{0.5,0.5,0.5}
\definecolor{codepurple}{rgb}{0.58,0,0.82}
\definecolor{backcolour}{rgb}{0.95,0.95,0.92}

\lstdefinelanguage{promptlanguage}{
    morecomment=[l][\color{codepurple}]{//},
    morecomment=[s][\color{blue}]{\{}{\}},
}

\lstdefinestyle{promptstyle}{
    backgroundcolor=\color{white},   
    commentstyle=\color{codegreen},
    keywordstyle=\color{magenta},
    numberstyle=\tiny\color{codegray},
    stringstyle=\color{codepurple},
    basicstyle=\ttfamily\small,
    frame = single,
    breakatwhitespace=false,         
    breaklines=true,                 
    captionpos=b,                    
    keepspaces=true,                 
    numbers=left,                    
    numbersep=5pt,                  
    showspaces=false,                
    showstringspaces=false,
    showtabs=false,                  
    tabsize=2
}

\newenvironment{chat} {
    
    \newcommand\who[1]{\par\hangindent=2.5em\hangafter=1 ##1:}
    \par\vskip2em
}{%
    \par
    \hangindent=0pt\hangafter=0
    \vskip2em
}

\title{Generating medically-accurate summaries of patient-provider dialogue: \\A multi-stage approach using large language models}

\author{%
  Varun Nair\\
  Curai Health\\
  \texttt{varun@curai.com}\\
  \And
  Elliot Schumacher\\
  Curai Health\\
  \texttt{elliot@curai.com}\\
  \And
  Anitha Kannan\\
  Curai Health\\
  \texttt{anitha@curai.com}\\
}

\begin{document}

\maketitle

\begin{abstract}
A medical provider's summary of a patient visit serves several critical purposes, including clinical decision-making, facilitating hand-offs between providers, and as a reference for the patient. An effective summary is required to be coherent and accurately capture all the medically relevant information in the dialogue, despite the complexity of patient-generated language. Even minor inaccuracies in visit summaries (for example, summarizing ``patient does not have a fever'' when a fever is present) can be detrimental to the outcome of care for the patient .

This paper tackles the problem of medical conversation summarization by discretizing the task into several smaller dialogue-understanding tasks that are sequentially built upon. First, we  identify medical entities and their affirmations within the conversation to serve as building blocks. We study dynamically constructing few-shot prompts for tasks by conditioning on relevant patient information and use GPT-3 \cite{brown2020gpt3} as the backbone for our experiments. We also develop GPT-derived summarization metrics to measure performance against reference summaries quantitatively. Both our human evaluation study and metrics for medical correctness show that summaries generated using this approach are clinically accurate and outperform the baseline approach of summarizing the dialog in a zero-shot, single-prompt setting.

\end{abstract}

\begin{figure}[t]
    \centering
    \includegraphics[width=0.48\textwidth]{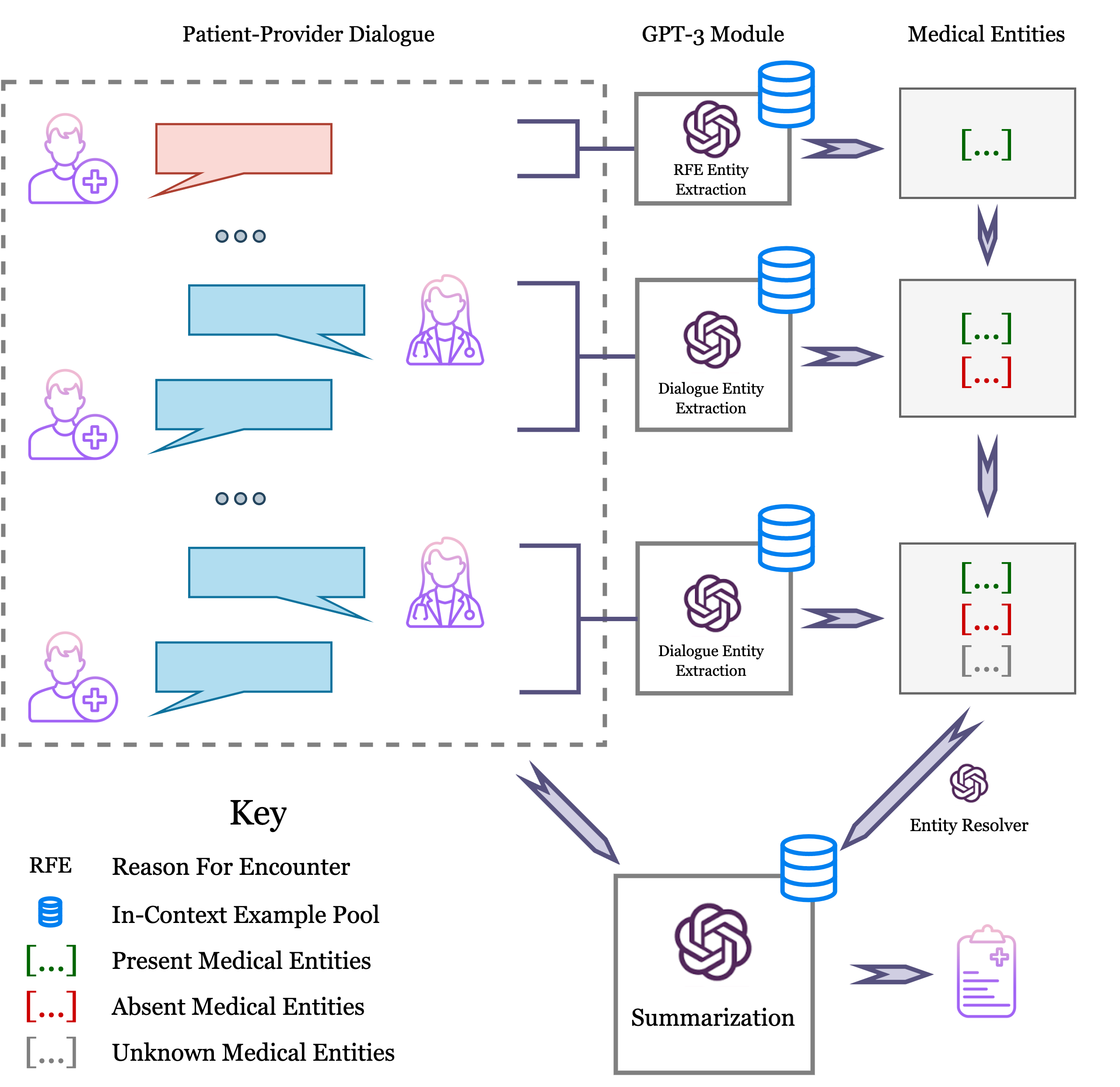}
    \caption{MEDSUM-ENT utilizes a multi-stage approach for medical dialogue summarization with GPT-3 that improves upon naive summarization. The approach utilizes intermediate model calls to extract medical concepts that inform summarization generation.}
    \label{fig:figure_1}
\end{figure}

\section{Introduction}
\label{sec:intro}
A critical clinical task during a medical encounter between a patient and a physician is summarizing the conversation. This summarized note, whether created by a physician or medical assistant, contains important information about the visit and serves as a reference for future patient visits and for the patient. Physicians often spend many hours each week performing such tasks.  Charting work, in general, has been identified as a contributing factor to increased rates of physician burnout \cite{eschenroeder2021Associations}.

Automating medical conversation summarization has been studied with limited success \cite{pivovarov-elhadad-2015-automated, liang-etal-2019-novel-system, gao-etal-2022-summarizing, macavaney2019ontology, chintagunta-etal-2021-medically}. Some methods try to directly summarize the chat \cite{enarvi-etal-2020-generating, zhang2021leveraging} while others pair deep learning methods with information extracted from knowledge bases to produce accurate summaries \cite{joshi-etal-2020-dr}. As base deep learning methods have improved and pre-trained language models specific to summarization such as PEGASUS \citep{zhang2020pegasus}, BART \citep{lewis-etal-2020-bart}, and GPT-3 \cite{brown2020gpt3} have emerged, we have seen increased fidelity of the summaries generated. However, performance is still not to a reliable standard in practical settings for several reasons. First, the lack of labeled clinical data makes it hard to build high-performance fine-tuned models. This reflects lower-than-expected specificity and accuracy in faithfully capturing medical concepts and their affirmations (e.g., present, absent, unknown). Second, custom-trained models need more world knowledge to understand patient language in these conversations and how they map to medical concepts. Third, these models often require breaking conversations into smaller segments to deal with limited context windows.  This in turn introduces challenges such as incorrect anaphora and coreference resolution across segmented pieces of the conversation.

The key contributions of this paper include;
\begin{itemize}
    \item {\bf MEDSUM-ENT}: Inspired by recent works of \citet{chintagunta-etal-2021-medically} and \citet{Agrawal2022LargeLM}, we introduce MEDSUM-ENT: a medical conversation summarization model that takes a multi-stage approach to summarization, using GPT-3 \cite{brown2020gpt3} as the backbone. MEDSUM-ENT ( Fig.~\ref{fig:figure_1}) grounds the task by first extracting medical entities and their affirmations. These extractions are included as additional input that informs the final summarization step through prompt chaining \citep{wu2022AIChains}. MEDSUM-ENT also exploits few-shot prompting for medical concept extraction and summarization through in-context example selection. 
    
    In both qualitative physician analysis of medical dialogue summaries and quantitative metrics, MEDSUM-ENT generates clinically accurate summaries and produces summaries that are preferable to a zero-shot, single prompt baseline.
    
    \item{\bf Automated metrics}: Quantitative metrics are hard to design for generative tasks. We extend proxy metrics of \citet{joshi-etal-2020-dr} by leveraging GPT-3 to compare the coverage of the presence of medical entities in the generated texts. Beyond only identifying exact matches, our approach better accounts for paraphrasing those medical events within the larger text.

\end{itemize}
%\import{sections/}{related_work.tex}

\section{Methods}
\label{sec:methods}

We now detail the components of our MEDSUM-ENT framework for medical dialogue summarization, represented in Figure \ref{fig:figure_1}. 

\paragraph{Medical Entity Extraction}
To highlight clinical concepts, we extract medical entities (symptoms, diseases etc.) and their affirmation status of either \textbf{present}, \textbf{absent}, or \textbf{unknown}. These entities and their status will be used as additional inputs to the final summarization step.

We first perform entity extraction on the patient's first message of the encounter, which is often lengthy and information dense. We call this message the \textit{reason for encounter} (RFE). Conversational turns between the medical provider and the patient follow the RFE.  We also extract medical entities from the conversation, one provider and one patient turn at a time. To accommodate these two types of texts, we use two different prompts, included in Appendix Prompt \ref{prompt:rfe} (for RFE entity extraction) and Appendix Prompt \ref{prompt:dialogue} (for dialogue entity extraction). Both prompts are populated with in-context examples (see In-Context Example Selection) along with the patient's age and sex. The final list of entities in the dialogue is obtained by collating all entities extracted across the RFE and all dialogue turns. 

Additionally, we also use an entity resolver similar to those used in \citet{Agrawal2022LargeLM} to resolve entities in the unknown entities list whose status may have changed over the course of the dialogue (see Appendix Prompt \ref{prompt:unknown_resolver}). For instance, a dialogue turn pair may not have enough information to definitively assign a present or absent status and is thus assigned an entity as "unknown".  A later dialogue turn may contain information that changes that assignment. By introducing this refinement step, we reduce mistakes in the "Pertinent Unknowns" section of the summary (see Table \ref{table:table_1}).

\paragraph{Summarization}
Given a list of medical entities, we summarize the medical dialogue using the dialogue and the entities as input. Our summaries are structured into six sections: \textit{Demographics and Social Determinants of Health}, \textit{Medical Intent}, \textit{Pertinent Positives}, \textit{Pertinent Negatives}, \textit{Pertinent Unknowns}, and \textit{Medical History} (see Appendix Prompt \ref{prompt:summarization} for details). 

\paragraph{In-Context Example Selection}
\label{sec:dynamic_prompting}

For the entity extraction and summarization modules, we compare semantic-similarity and random in-context example selection. Semantic-similarity-based selection selects labeled examples from a pool using the patient's age, sex, and the query point. Random selection randomly selects in-context examples from these pools to populate our prompts. Further implementation details are in Appendix \ref{app:dynamic}.

\begin{figure}[!]
    \centering
    \includegraphics[height=0.45\textwidth]{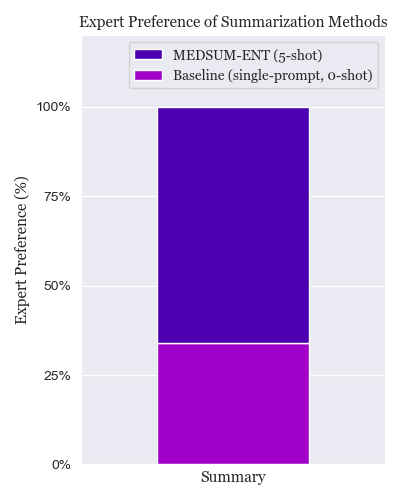}
    \caption{Results of human expert evaluations show MEDSUM-ENT (5-shot) is preferred 66\% to 34\% over a single-prompt, 0-shot naive summarization baseline.}
    \label{fig:figure_2}
\end{figure}

\begin{table*}[t!]
    \centering
    \resizebox{1\textwidth}{!}
    {
%\begin{tabular}{cccccccccc}
\begin{tabular}{cP{1.5cm}P{2.2cm}P{1.5cm}P{1.5cm}|P{1.5cm}P{1.8cm}P{1.8cm}P{1.5cm}P{1.5cm}}
\toprule
 &  & & &  & \multicolumn{5}{c}{GPT-F1 ($\uparrow$)} \\
Method & Extraction K-shot & Summarization K-shot & Example Selection & Entity Resolver & Pertinent Positives & Pertinent Negatives & Pertinent Unknowns & Medical History & Average\\ \hline
\multirow{3}{*}{Naive} & - & 0-shot & - & - & \textbf{72.9} & \textbf{71.7} & 45.4 & 43.9 & 58.5 \\
 & - & 1-shot & semantic & - & 71.0 & 69.5 & 42.1 & 48.3 & 57.7 \\
 & - & 1-shot & random & - & 69.4 & 69.1 & 47.5 & 44.7 & 57.7 \\\hline
\multirow{7}{*}{MEDSUM-ENT} & 1-shot & 1-shot & semantic & \checkmark & 72.4 & 70.1 & 50.0 & 46.2 & 59.7 \\
 & 1-shot & 1-shot & random & \checkmark & 71.4 & 71.1 & 54.0 & 48.3 & 61.2 \\
 & 3-shot & 1-shot & semantic & \checkmark & 71.9 & 69.0 & 42.5 & 47.0 & 57.6 \\
 & 3-shot & 1-shot & random & - & 72.1 & 69.4 & 46.4 & 45.8 & 58.4 \\
  & 3-shot & 1-shot & random & \checkmark & 72.2 & 70.9 & \textbf{55.8} & \textbf{50.4} & \textbf{62.3} \\
 & 5-shot & 1-shot & semantic & \checkmark & 71.8 & 70.2 & 46.6 & 46.3 & 58.7 \\
 & 5-shot & 1-shot & random & \checkmark & 71.9 & 68.3 & 51.9 & 48.2 & 60.0 \\ \bottomrule
\end{tabular}
}
    \caption{Results of GPT-driven metrics. Performance across ``Pertinent Positives'', ``Pertinent Negatives'' sections are fairly consistent across methods. MEDSUM-ENT demonstrates consistently improved performance in the ``Pertinent Unknowns'' and ``Medical History'' sections. Surprisingly, we also find consistently higher performance across experiments using random in-context example selection over semantic-similarity-based selection.
}
    \label{table:table_1}
\end{table*}

\section{Experiments}
\label{sec:exps}

\paragraph{Dataset:} We use a dataset of 100 clinical encounters of dialogue-summary pairs that occurred between a licensed physician and a patient on a telehealth platform. Encounters in this dataset cover a wide variety of common presentations in telehealth, including urinary tract infections, back/abdominal pains, toothaches, and others. All data was de-identified and scrubbed for protected health information prior to experimentation. Conversations contain 46 dialogue turns on average (min of 8 turns, max of 92 turns) and an average of 2342 unigram tokens per encounter.  Ground truth summaries were created by using \texttt{text-davinci-002} on encounter data to generate an initial summary, which physicians then edited for correctness.

\paragraph{Baselines/Ablations:} 
We compare MEDSUM-ENT to a ``naive'' zero-shot, single-prompt baseline (i.e. without chaining) that prompts GPT-3 to summarize the conversation (see Appendix Prompt \ref{prompt:baseline_summarization}). For MEDSUM-ENT, we evaluate extraction k-shot configurations (1,3,5-shot) and in-context example selection methods (semantic-similarity based, random) for entity extraction. We use RFE and dialogue entity extraction prompts in at least a 1-shot configuration for MEDSUM-ENT to ensure valid output and formatting. Our summarization prompt for baselines and MEDSUM-ENT cannot go beyond 1-shot due to token limit constraints. All experiments are run once and leverage GPT-3 (davinci-003) for generation (see Appendix \ref{app:experimentdetails} for temperature, max\_tokens, and top\_p settings for each prompt). 

\subsection{Evaluation Metrics}

\paragraph{Expert Evaluation}
We also asked four doctors, who serve patients on a telehealth platform, to judge between the MEDSUM-ENT and baseline-generated summaries on three points on a random set of 50 encounters. For a given encounter, we asked 1) for preference between baseline and MEDSUM-ENT summaries, 2) what amount of clinical information was captured in MEDSUM-ENT's summaries, and 3) about the presence of clinically harmful information in MEDSUM-ENT summaries (see Appendix \ref{app:expert} for exact instructions and other details).

\paragraph{GPT-Driven Automated Summarization Metrics:}\label{sec:GPT-metric} Acknowledging the challenges in automatic evaluations of summarization \cite{peyrard-2019-studying, goyal2022news}, we focus on quantitatively evaluating the correctness/faithfulness of capturing medical concepts and their affirmation status. 

We extend the approach to metrics in \citet{joshi-etal-2020-dr} to have two components, both powered by GPT-3: a medical concept extractor (Appendix Prompt \ref{prompt:extract}) and a verifier (Appendix Prompt \ref{prompt:verify_concepts}). The verifier checks if the concepts extracted from one piece of text are present in another and permits the same medical concept extracted or written in different ways to count towards a true positive. For example, for the ``Pertinent Positives'' section, the predicted value may be \textit{``Patient has back pain and COVID-19''} and the resulting concepts [``back pain'', ``COVID-19''] and the ground-truth \textit{``Patient has COVID and some pain in the backside''} with concepts [``COVID'', ``pain in the back'']. Prior metrics that rely on verbatim matches would fail to recognize the predicted text as correct. We define the following metrics: 

\textbf{GPT-Recall}: We extract medical entities from both the predicted text and ground-truth text of the same summary section. We use the verifier to infer if the entities extracted from the ground-truth section are also present in the predicted text. This produces $tp_{gt}$ and $f_{n}$, which is used to calculate $\textnormal{GPT-Recall} = \frac{tp_{gt}}{tp_{gt}+f_{n}}$.

\textbf{GPT-Precision}: We verify concepts extracted from the predicted section are also present in the ground-truth text, either as exact matches or re-phrasings. This produces $tp_{pred}$ and $f_{p}$, which is used to calculate $\textnormal{GPT-Precision} = \frac{tp_{pred}}{tp_{pred} + f_{p}}$.

\textbf{GPT-F1} is the harmonic mean of GPT-Precision and GPT-Recall. Note our approach maintains the integrity of recall and precision (neither score can take on a value > 1). We evaluate MEDSUM-ENT via the GPT-Precision and GPT-Recall metrics described in section \ref{sec:GPT-metric} on all 100 clinical encounters.

\section{Results} % (Varun)

Table \ref{table:table_1} shows quantitative metrics on summaries produced by the baselines and MEDSUM-ENT.  Both generated summaries are compared to the ground truth summaries. We see that while GPT-F1 performance for ``Pertinent Positives'' and ``Pertinent Negatives'' is consistent across methods, MEDSUM-ENT's ability to capture the ``Pertinent Unknowns'' and ``Medical History'' pushes its average consistently above that of the naive zero-shot, non-chained baseline. These sections are crucial to include correctly as they often influence clinical decision-making.  Also, the Unknown Entity Resolver improves performance specifically in the ``Pertinent Unknowns'' section (ablated in rows 7 vs. 8  with 46.4 vs. 55.8 for with and without the resolver). The ``Demographics and Social Determinants of Health'' and ``Medical Intent'' sections have nearly identical, accurate output across all experiments, so we do not calculate metrics for them. See Appendix \ref{app:qualitative} for example generated summaries.

We find two surprising results.  First, there is no correlation between a larger k-shot and increased performance.  This may demonstrate diminishing returns of GPT-3 to perform medical concept extraction. Furthermore, the use of semantic similarity to select in-context examples performs \textbf{worse} than randomly selecting examples. This follows \citet{Ye2022ComplementaryEF} which claims diversity of in-context samples is more important than similarity. 

In our expert human evaluations, Figure \ref{fig:figure_2} demonstrates MEDSUM-ENT (5-shot, semantic) summaries are preferred over the baseline summaries 66\% to 34\%. Our expert evaluators also rate MEDSUM-ENT capturing all relevant medical information in 40\% of evaluated summaries, most information in 48\%, some information in 12\%, and zero information in 0\%. This provides further qualitative evidence for MEDSUM-ENT's ability to effectively summarize. However, our expert evaluators also rate 28\% of the summaries evaluated as containing incorrect information that could harm the patient if acted on by medical providers. Often these are due to misattributed symptoms and conditions (e.g., symptoms marked as absent but were present, missed medication allergies). This is consistent with the GPT-F1 measures for pertinent positives and negatives in Table \ref{table:table_1} and highlights the challenge involved in deploying a system such as MEDSUM-ENT. Further work is needed to trust such systems in the wild.
\vspace{0em}
%While encouraging, this highlights that MEDSUM still suffers from failure modes that occasionally made our expert evaluators prefer the naive summary. Comments left by our evaluators suggest that additional modules for resolving pertinent positives, negatives, and medical history sections may benefit performance.

%Lastly, it is also important to note that these GPT-based metrics are not perfect, and further work in evaluating GPT-generated summaries is needed.

\section{Conclusion}
In this work, we introduce MEDSUM-ENT, a multi-stage framework for medical dialogue summarization that modularizes summarization into multiple stages that extract and refine medical entities from dialogue turns. Through human evaluation and quantitative metrics, we show that this method is clinically accurate and preferable to naive zero-shot summarization with GPT-3. We hope that future work can investigate refinement modules and iterative summarization further and conduct wider expert human evaluation studies to better understand the challenges in bringing model-assisted summarization to medical providers in the near term.

\section{Limitations}
The experiments in this paper were performed using OpenAI's GPT-3 API.  While running locally does not require a large amount of computational resources, the server-side service cannot be easily replicated and requires a large amount of computational resources.  Additionally, given the inherently restrictive nature of medical text, we can only evaluate our approach on a small corpus of English-language dialogues taken from the dataset of a single company's medical service, which we cannot release due to privacy concerns.  Finally, given summarization is a challenging task to evaluate, we rely on a small number of expert human annotators and automatic metrics. However, additional annotations may be helpful and it may also help to study and report labeler agreement when reporting human preferences.
%Note: limitations must be last, and does not count towards the page limit.

\bibliography{main,anthology}
\bibliographystyle{acl_natbib}

\appendix

\newpage
\section{Appendix}
\subsection{Dynamic example selection}\label{app:dynamic}
We create labeled in-context example pools for RFE entity extraction and dialogue entity extraction using physician labels for what medical concepts would have been extracted and created a summarization pool using physician-written dialogue summaries. The dialogue summaries for this pool were created by physicians editing the outputs of summaries created by text-davinci-002. Semantic-similarity based example selection is implemented using nearest-neighbor search with the LangChain\footnote{https://github.com/hwchase17/langchain} and FAISS \citep{johnson2019billion} libraries. 

\subsection{Experiment details}\label{app:experimentdetails}
\begin{table}[h]
    \centering
    \resizebox{0.48\textwidth}{!}{
    \begin{tabular}{c|c c c}
    \toprule
        Prompt & temperature & max\_tokens & top\_p \\
        \hline
        RFE Medical Entity Extr. & 0.1 & 200 & 1.0 \\
        Dialogue Medical Entity Extr. & 0.1 & 200 & 1.0\\
        Unknown Entity Resolver & 0.1 & 200 & 1.0\\
        Summarization & 0.7 & 512 & 1.0\\
        Metric: Medical Entity Extr. & 0.0 & 200 & 1.0\\
        Metric: Medical Entity Verif. & 0.0 & 200 & 1.0\\
        \bottomrule
    \end{tabular}
    }
    \caption{Experimental settings for all prompts used in this work, no hyper-parameter search was run to obtain these values. We use lower temperature values for model calls where we expect lower variability in its inputs (summarization takes in dialogues and list of medical entities of varying lengths and sizes respectively, thus has a higher temperature). Running the metric concept extraction and verification prompts at a temperature of 0 ensures maximal reproducibility of metric computation. Each experiment (line in Table \ref{table:table_1} took approximately 3 hours to run, with exponential back-off used during GPT-3 queries.)}
    \label{tab:experimentdetails}
\end{table}

\subsection{Expert evaluation}\label{app:expert}
To qualitatively evaluate our summaries, we conducted physician evaluations focused on three questions: 

\begin{itemize}
    \item Q1: \textit{How often are summaries written using MEDSUM preferred over naively generated summaries?}
    \item Q2: \textit{What fraction of relevant clinical information is captured in the summaries generated by our method? (All, Most, Some, None)}
    \item Q3: \textit{Does the summary generated by our method contain incorrect information that could significantly alter the course of treatment and potentially harm the patient if this summary was used by another medical provider?}
\end{itemize}

Q1 was asked alongside some basic instructions for how choices should be made, shown below:\newpage
\noindent\rule{0.48\textwidth}{1pt}
 \vspace{1em}
For this task, you will see a dialogue, RFE, and age/sex information for a patient. The task is to identify which of the 3 summary options you would prefer to use as a visit summary.
You may use your own discretion in selecting which of the 3 options you prefer. Some things to note when selecting are:
\begin{itemize}
    \item How thorough and clinically accurate is each summary?
    \item Is the summary missing clinically relevant information?
    \item Does the summary contain extraneous information that is harmful if a provider were to read and act upon information in the summary?
    \item Which summary is stylistically preferable and/or easier to read?
\end{itemize}
\noindent\rule{0.48\textwidth}{1pt}

\vspace{3em}
\clearpage
\subsection{Qualitative Analysis}
\label{app:qualitative}
We provide two examples of outputs from our naive 0-shot, single-prompt baseline and MEDSUM (5-shot, semantic) below. All protected health information is anonymized using tags in all caps (eg. DATE\_1, LOCATION\_1, NAME). Note the green text which captures correctly summarized information that is not present in the other method's output, and red text which marks a mistake that is not present in the other method's output.
\noindent\rule{0.48\textwidth}{1pt}
\textit{\textbf{Chat A}}.  \textit{Reason for Encounter:} UTI
\vspace{-1em}
\begin{chat}
\who{Doctor} Do you have any pain or discomfort when you urinate ? 
\who{Patient} present  
\who{Doctor} Do you experience sudden urges to urinate ? 
\who{Patient} present  
\who{Doctor} Have you been urinating more frequently than usual ? 
\who{Patient} present  
\who{Doctor} Have you noticed any discharge from the urethra ? 
\who{Patient} present  
\who{Doctor} Have you been sexually active in the past 3 months ? 
\who{Patient} present  
\who{Doctor} Do you have to urinate at night , waking up two or more times to urinate ? 
\who{Patient} present  
\who{Doctor} Do you have pain on the side of your torso just below your ribs ? 
\who{Patient} absent  
\who{Doctor} Does your abdomen or belly hurt ? 
\who{Patient} unknown  
\who{Doctor} Since when are you having these symptoms ? 
\who{Patient} 2 weeks ago 
\who{Doctor} Have you taken anything for these symptoms ? 
\who{Patient} Azo 
\who{Doctor} Has it helped ? 
\who{Patient} Somewhat not much 
\who{Doctor} Can you describe the urine ? 
\who{Patient} Lots of it Bloated Orange when I take that medicine Dark yellow the rest 
\who{Doctor} Is there any blood in the urine ? 
\who{Patient} No 
\who{Doctor} Does the urine have any foul smell ? 
\who{Patient} No 
\who{Doctor} Do you have pain below your belly button ? 
\who{Patient} present  Some not much 
\who{Doctor} Have you noticed any difficulty passing urine ? 
\who{Patient} absent  
\who{Doctor} Do you have pain with sexual intercourse ? 
\who{Patient} present  This is when it started 
\who{Doctor} Do you have pain in lower abdomen lower belly ? 
\who{Patient} present  
\who{Doctor} Have you noticed any vaginal discharge ? 
\who{Patient} unknown  
\who{Doctor} Do you have low back pain ? 
\who{Patient} present  
\who{Doctor} Does the outer part of your vagina itch ? 
\who{Patient} absent  
\who{Doctor} Do you have a fever ? 
\who{Patient} absent  
\who{Doctor} Do you have nausea ? 
\who{Patient} present  
\who{Doctor} What was the first day of your last menstrual period ? 
\who{Patient}  I have no cycle hysterectomy 
\who{Doctor} When were you last sexually active ? 
\who{Patient}  Two weeks ago 
\who{Doctor} Have you had a UTI in the past ? 
\who{Patient} Yes 
\who{Doctor} When was it [NAME] and what was done ? 
\who{Patient} Six months ago antibiotics
\end{chat}
\noindent\rule{0.48\textwidth}{1pt}
\clearpage

\begin{minipage}[t]{.46\textwidth}
        \centering
        \textbf{MEDSUM Summary for Chat A}
    \begin{tcolorbox}[width=.9\textwidth,
                  boxsep=0pt,
                  arc=0pt,
                  colback=white,
                  left=2pt,
                  right=2pt,
                  top=2pt,
                  bottom=2pt
                  ]%%
                  \textbf{Demographics and Social Determinants of Health:} \\
 A 46 year old female. \\\\
 \textbf{Patient Intent:} \\
 Patient came for UTI. \\\\
 \textbf{Pertinent Positives:} \\
 Patient reports pain and discomfort when urinating, sudden urges to urinate, increased urination frequency, urethral discharge, sexual activity in the past 3 months, orange urine when taking medication, dark yellow urine the rest, abdominal pain below the belly button, pain with sexual intercourse when the symptoms started, lower abdominal pain, low back pain and nausea. {\color{codegreen} \textbf{Symptoms have been present for 2 weeks.}} She has taken Azo to manage the symptoms which have only slightly helped. 
\\\\
\textbf{Pertinent Unknowns}: \\
 Patient is unsure if her urine has any foul smell and if there is any vaginal discharge. 
 \\\\
 \textbf{Pertinent Negatives}: \\
 Patient reports no fever, no difficulty passing urine, no vulvar pruritus, and no hematuria.  
\\\\
\textbf{Medical History}: 
 Patient has a hysterectomy and was last sexually active two weeks ago. She had a UTI six months ago and was treated with antibiotics.
    \end{tcolorbox}

    \end{minipage}%
        \begin{minipage}[t]{.48\textwidth}
\textbf{Baseline Summary for Chat A}
                \centering
\begin{tcolorbox}[width=.9\textwidth,
                  boxsep=0pt,
                  arc=0pt,
                  colback=white,
                  left=2pt,
                  right=2pt,
                  top=2pt,
                  bottom=2pt
                  ]%%
                \textbf{Demographics and Social Determinants of Health:} \\
 The patient is a 46 year old female. 
 \\

 \textbf{Patient Intent:}\\
 The patient is seeking care for UTI.
 \\

\textbf{Pertinent Positives:}\\
 The patient has experienced pain or discomfort when urinating, sudden urges to urinate, increased urination frequency, discharge from the urethra, and pain with sexual intercourse in the past 3 months. She has also experienced pain on the side of her torso below her ribs, and pain in her lower abdomen. She has taken Azo, which somewhat helped her symptoms. The urine is abundant and bloated, orange when she takes the medicine, and dark yellow the rest of the time.
 \\

\textbf{Pertinent Unknowns:} \\
 The patient is unsure of {\color{red} \textbf{any abdominal pain}}, and if there is any foul smell to the urine. She is also unsure if there is any vaginal discharge.
 \\

\textbf{Pertinent Negatives: }\\
 The patient does not have any blood in the urine, and the outer part of her vagina does not itch. She also does not have a fever or difficulty passing urine.
 \\

\textbf{Medical History: }\\
 The patient has had a hysterectomy and was last sexually active two weeks ago. She has had a UTI in the past, which was six months ago, and she was treated with antibiotics.
                  
    \end{tcolorbox}
    \end{minipage}
\clearpage
\noindent\rule{0.48\textwidth}{1pt}
\textit{\textbf{Chat B}}.  \textit{Reason for Encounter:} I took my inhaler a few minutes and realized i am almost out of it. I was woundering if i can get a prescription from you? I have asthma and the last few days tight chest and breath with weezing issues
\vspace{-1em}
\begin{chat}
\who{Doctor} Thanks for confirming Sorry to hear about the symptoms you are currently experiencing , [NAME] . May I know when you were diagnosed with asthma ?
\who{Patient} When i was born I was dx Last asthma attack has been at least 2 years
\who{Doctor} Thanks for letting me know about it . Which inhaler have you been prescribed on and how long have you been using it ?
\who{Patient} Albuteraol
\who{Doctor} Okay . Would you be able to share a picture of the inhaler with the last prescription , that clearly mentions about the dosage and frequency ?
\who{Patient} Prn . I do not have the prescription . I got it almost 2 years ago
\who{Doctor} Thanks for sharing , [NAME] . Have you been using it since childhood ?
\who{Patient} I have been using this one only on prn bases . In the past i have used a steroid one spary twice a day . Do not remember the name
\who{Doctor} Okay . How often do you generally use the inhaler and how many puffs do you use each time ?
\who{Patient} Albuterol i use maybe best guess once a month too once every two months . When i use it two sparys . I take one wait for 5 minutes and repeat
\who{Doctor} Okay . When was the last \who{Doctor}s visit ?
\who{Patient} In January . Just lost my job so i dont have insurance to go back right now
\who{Doctor} I hear you , [NAME] . Sorry to hear about the job loss . How long have you been experiencing these symptoms now ? . Have you noticed any trigger factors associated with them ? Anything that could have resulted in the symptoms or worsened it ?
\who{Patient} Possible since i am obsessed 340 lbs . But the weather here has been muggy hot humidity has been off the charts . Not normal for LOCATION\_1
\who{Doctor} Thanks for letting me know about it . Im going to send you questions to make sure Im correctly collecting all of your symptoms . Please select “Yes” , “No” , or “Unsure” in the question and hit Send once youve input your response Do you get more short of breath than expected with activity ?
\who{Patient} present  Lately yes
\who{Doctor} Do you have a dry cough ?
\who{Patient} present  Chronic
\who{Doctor} Do you have a cough that brings up phlegm or mucus ?
\who{Patient} present  Clear
\who{Doctor} Do you have chest pain that worsens when you take a deep breath ?
\who{Patient} absent 
\who{Doctor} Do you wake up from sleep and find yourself out of breath ?
\who{Patient} absent 
\who{Doctor} Do you feel short of breath while lying down ?
\who{Patient} absent 
\who{Doctor} Do you have shortness of breath at rest or while sitting ?
\who{Patient} absent 
\who{Doctor} Do you have difficulty falling asleep ?
\who{Patient} absent 
\who{Doctor} Do you frequently find yourself waking up in the middle of the night ?
\who{Patient} present 
\who{Doctor} Do you feel fatigued ?
\who{Patient} present  Possible due to the over weight that i am currently working on loosing
\who{Doctor} Have you recently been exposed to animals ?
\who{Patient} absent 
\who{Doctor} Have you recently been in close contact with someone diagnosed with COVID-19 ?
\who{Patient} absent 
\who{Doctor} Do you have difficulty breathing when you lean forward ?
\who{Patient} present 
\who{Doctor} Do you have difficulty speaking more than short phrases due to the breathing difficulty ?
\who{Patient} absent 
\who{Doctor} How long have you been noticing all of these symptoms , [NAME] ?
\who{Patient} Short of breath leaning forward past 30 minutes . Took two puffs starting to feel ok NAME\_1... was dx with sinus issues . Always draining
\who{Doctor} Has the breathing difficulty , wheezing , chest pressure started only today within the past hour ?
\who{Patient} Wheezing yesterday . Chest pressure and wheezing only today

\end{chat}
\noindent\rule{0.48\textwidth}{1pt}

\clearpage
\begin{minipage}[t]{.46\textwidth}
        \centering
        \textbf{MEDSUM Summary for Chat B}
    \begin{tcolorbox}[width=.9\textwidth,
                  boxsep=0pt,
                  arc=0pt,
                  colback=white,
                  left=2pt,
                  right=2pt,
                  top=2pt,
                  bottom=2pt
                  ]%%
\textbf{Demographics and Social Determinants of Health:} \\
 A 43 year old male. He recently lost his job and currently lacks insurance. He is currently working on losing weight and the weather in LOCATION\_1 has been extremely muggy and hot. \\

\textbf{Patient Intent: } \\
 Patient came to address his concern of needing a prescription for his inhaler and to address his asthma symptoms of tight chest, shortness of breath, and wheezing.\\

\textbf{Pertinent Positives: }\\
 Patient has asthma since birth, he has been using an Albuterol inhaler as needed and a steroid spray twice a day in the past, he has a dry cough, difficulty breathing when leaning forward, is coughing up phlegm, {\color{codegreen} \textbf{has chest pressure, wakes up in the middle of the night, is fatigued,}} has sinus issues, and has wheezing. \\

\textbf{Pertinent Unknowns: }\\
 Patient is unsure of the dosage and frequency of his prescription medication and the name of his steroid medication.\\

\textbf{Pertinent Negatives:}\\ 
 Patient has no chest pain, {\color{codegreen} \textbf{difficulty speaking more than short phrases because of breathing difficulty},} shortness of breath while lying down, difficulty initiating sleep, mucus, {\color{codegreen} \textbf{animal exposure, or contact with someone diagnosed with COVID19}}. \\

\textbf{Medical History: }\\
 Patient was diagnosed with asthma at birth, his last asthma attack was at least 2 years ago, he has been using an Albuterol inhaler as needed for the past 2 years, and he has been experiencing symptoms for the past 30 minutes.
    \end{tcolorbox}

    \end{minipage}%
        \begin{minipage}[t]{.48\textwidth}
\textbf{Baseline Summary for Chat B}
                \centering
\begin{tcolorbox}[width=.9\textwidth,
                  boxsep=0pt,
                  arc=0pt,
                  colback=white,
                  left=2pt,
                  right=2pt,
                  top=2pt,
                  bottom=2pt
                  ]%%
 \textbf{Demographics and Social Determinants of Health:}\\
 43 year old male patient, unemployed and obese (340 lbs). Living in a hot, humid area with high levels of humidity.\\

\textbf{Patient Intent:}\\
Patient is requesting a prescription for an inhaler, as the patient is almost out of it and has been experiencing tightness in their chest and wheezing issues for the past few days.\\

\textbf{Pertinent Positives:}\\
Patient has asthma, has been using Albuterol inhaler on a prn basis, has a chronic dry cough, and has been noticing shortness of breath and difficulty leaning forward for the past 30 minutes. \\

\textbf{Pertinent Unknowns:}\\
{\color{codegreen} \textbf{Patient is unsure if the weather or being overweight is causing the symptoms.}}\\

\textbf{Pertinent Negatives:}\\
Patient does not have chest pain, difficulty breathing when lying down, difficulty speaking, difficulty falling asleep, or {\color{red} \textbf{waking up in the middle of the night due to breathing difficulty.}}\\

\textbf{Medical History:}\\
Patient was diagnosed with asthma at birth and had their last asthma attack 2 years ago. Patient was last at the doctor in January and has been experiencing the symptoms for the past few days. Patient has been using the current inhaler for 2 years, and has used a steroid inhaler twice a day in the past, but can not remember the name. Patient has been diagnosed with sinus issues and has been having clear mucus drainage.
                  
\end{tcolorbox}
\end{minipage}
\clearpage

\begin{lstlisting}[language=promptlanguage,float=*,caption=Prompt for reason for encounter (RFE) medical entity extraction.,label={prompt:rfe}]
----
Below is the first message from a{age_and_sex} patient seeking care:
----
Patient: {rfe}
----
Using the patient's message above, please find the medical entities (medical concepts, symptoms, or medical conditions) and each one's status (present, absent, or unknown) that would be important for a doctor to know.
If the patient states the presence of a medical concept, symptom or condition, the medical entity's status should be present. 
If the patient denies the presence of a medical concept, symptom or condition, the medical entity's status should be absent.
Medical entities should have an unknown status ONLY if the patient is themselves unsure or hesitant about a medical entity (eg: an answer of "Unknown", or "I'm not sure about..."").
Do NOT add a medical entity as unknown if the uncertainty is due to a DATE_1, DATE_2, NAME, or LOCATION tag. If there is such a medical entity associated with a tag, it must be either positive or negative.
Only extract medical entities that exist in the patient's message. DO NOT EXTRACT NON-MEDICAL ENTITIES.
Each medical entity should belong to one of six categories: Demographics and Social Determinants of Health, Patient Intent, Pertinent Positives, Pertinent Negatives, Pertinent Unknowns, or Medical History.
----
\end{lstlisting}

\begin{lstlisting}[language=promptlanguage,float=*,caption=Prompt for dialogue medical entity extraction.,label={prompt:dialogue}]
----
Below is a dialogue between a doctor and a{age_and_sex} patient seeking care:
----
{dialogue}
----
Using the patient's message above, please find the medical entities (medical concepts, symptoms, or medical conditions) and each one's status (present, absent, or unknown) that would be important for a doctor to know.
If the patient states the presence of a medical concept, symptom or condition, the medical entity's status should be present. 
If the patient denies the presence of a medical concept, symptom or condition, the medical entity's status should be absent.
Medical entities should have an unknown status ONLY if the patient is themselves unsure or hesitant about a medical entity (eg: an answer of "Unknown", or "I'm not sure about..."").
Do NOT add a medical entity as unknown if the uncertainty is due to a DATE_1, DATE_2, NAME, or LOCATION tag. If there is such a medical entity associated with a tag, it must be either positive or negative.
Only extract medical entities that exist in the patient-physician dialogue. DO NOT EXTRACT NON-MEDICAL ENTITIES.
Each medical entity should belong to one of six categories: Demographics and Social Determinants of Health, Patient Intent, Pertinent Positives, Pertinent Negatives, Pertinent Unknowns, or Medical History.
----
\end{lstlisting}

\begin{lstlisting}[language=promptlanguage,float=*,caption=Prompt for resolving unknown entities.,label={prompt:unknown_resolver}]
---
Below are medical entities (concepts, symptoms, or conditions) extracted from a medical conversation between a 22 year old patient and doctor.

Your job is to clean up the "Unknown Entities" list given the patient-doctor dialogue and a list of positive and negative entities. 

Remove any entities that that not are medical entities, or any entities that are unnecessary or completely irrelevant entities given the dialogue, positive entities, and negative entities. 

If a similar entity is both present and unknown, or, both absent and unknown, keep it in the unknowns ONLY if this information is still unknown after the entire dialogue.

---Dialogue---
//Example dialogue
---Dialogue---

Positive Entities: cough, headache, lower back pain

Negative Entities: fever, chest pain, chest tightness

Unknown Entities: past episode of flu, age_1, covid vaccination, symptoms, cough, fever, difficulty breathing, runny nose, frequency of headache, headache
---
Cleaned Unknown Entities: past episode of flu, covid vaccination, difficulty breathing, runny nose
---
---
Below are medical entities (concepts, symptoms, or conditions) extracted from a medical conversation between a {age_and_sex} patient and doctor.

Your job is to clean up the "Unknown Entities" list given the patient-doctor dialogue and a list of positive and negative entities. 

Remove any entities that that not are medical entities, or any entities that are unnecessary or completely irrelevant entities given the dialogue, positive entities, and negative entities. 

If a similar entity is both present and unknown, or, both absent and unknown, keep it in the unknowns ONLY if this information is still unknown after the entire dialogue.

---Dialogue---
Patient: {rfe}
{dialogue}
---Dialogue---

Positive Entities: {positive_entities}

Negative Entities: {negative_entities}

Unknown Entities : {unknown_entities}
---
Cleaned Unknown Entities:
\end{lstlisting}

\begin{lstlisting}[language=promptlanguage,float=*,caption=Prompt for MEDSUM summarization.,label={prompt:summarization}]
Below is a medical encounter between a {age_and_sex} patient and a doctor done over chat. 
The reason for the visit is: "{rfe}".
----
Medical Encounter #2
----
Patient: {rfe}
{dialogue}
----
Below are the medical entities and their status extracted from the patient-doctor dialogue from medical encounter #2. These entities can be used to help summarize the conversation below, but must be placed in the correct section (Demographics and Social Determinants of Health, Patient Intent, Pertinent Positives, Pertinent Unknowns, Pertinent Negatives, Medical History).

Positive Entities: {positive_entities}

Negative Entities: {negative_entities}

Unsure Entities: {unknown_entities}
----
Summary Instructions
----
Provide a summary of the medical encounter #2 between the doctor and the {age_and_sex} patient in 6 sections (Demographics and Social Determinants of Health, Patient Intent, Pertinent Positives, Pertinent Unknowns, Pertinent Negatives, Medical History).

Use the extracted entities to help summarize and place them in the appropriate section. Medical entities can be appropriate for any of the 6 sections and should be presented in an organized fashion. 

Add any important details from the dialogue to further explain, elaborate, or qualify a medical entity. If a medical entity is clinically inaccurate or completely irrelevant to the summary of the encounter, then do not summarize it.

The 6 sections to write the summary with are Demographics and Social Determinants of Health, Patient Intent, Pertinent Positives, Pertinent Unknowns, Pertinent Negatives, and Medical History. The definitions of each section are listed below.

Demographics and Social Determinants of Health: 
//Definition of section

Patient Intent: 
//Definition of section

Pertinent Positives: 
//Definition of section

Pertinent Unknowns:
//Definition of section

Pertinent Negatives: 
//Definition of section

Medical History: 
//Definition of section

----
Summary of Medical Encounter #2
----
\end{lstlisting}

\begin{lstlisting}[language=promptlanguage,float=*,caption=Prompt for naive zero-shot single-prompt summarization.,label={prompt:baseline_summarization}]
Below is a medical encounter between a {age_and_sex} patient and a doctor done over chat. 
The reason for the visit is: "{rfe}".
----
Medical Encounter
----
Patient: {rfe}
{dialogue}
----
Summary Instructions
----
Provide a summary of the medical encounter between the doctor and the {age_and_sex} patient in 6 sections (Demographics and Social Determinants of Health, Patient Intent, Pertinent Positives, Pertinent Unknowns, Pertinent Negatives, Medical History).

Use the extracted entities to help summarize and place them in the appropriate section. Medical entities can be appropriate for any of the 6 sections and should be presented in an organized fashion. 

Add any important details from the dialogue to further explain, elaborate, or qualify a medical entity. If a medical entity is clinically inaccurate or completely irrelevant to the summary of the encounter, then do not summarize it.

The 6 sections to write the summary with are Demographics and Social Determinants of Health, Patient Intent, Pertinent Positives, Pertinent Unknowns, Pertinent Negatives, and Medical History. The definitions of each section are listed below.

Demographics and Social Determinants of Health: 
//Definition of section

Patient Intent: 
//Definition of section

Pertinent Positives: 
//Definition of section

Pertinent Unknowns:
//Definition of section

Pertinent Negatives: 
//Definition of section

Medical History: 
//Definition of section

----
Summary of Medical Encounter
----
\end{lstlisting}

\begin{lstlisting}[language=promptlanguage,float=*,caption=Prompt for extracting medical concepts in metric computation.,label={prompt:extract}]
Given the following snippet of a medical dialogue summary, extract the medical concepts (symptoms, diseases, conditions, allergies, lab tests, etc.) present.

The heading of the section from which the summary was extracted will also be provided.

---Example 1---
Pertinent Negatives:  Patient reports no <concept_1>, no <concept_2>, <concept_3>, and <concept_4>. Patient also reports having no trouble with <concept_5>.

Medical Concepts: [<concept_1>, <concept_2>, <concept_3>, <concept_4>, <concept_5>]
---Example 1---

---Example 2---
Pertinent Positives:  Patient ongoing <concept_1> for the past 5 days, <concept_2>, and some <concept_3>. Patient had <concept_4> done in May 2021.

Medical Concepts: [<concept_1>, <concept_2>, <concept_3>, <concept_4>]
---Example 2---

---Example 3---
Pertinent Unknowns:  Patient is unsure about <concept_1> and <concept_2>.

Medical Concepts: [<concept_1>, <concept_2>]
---Example 3---

---Example 4---
Medical History: Patient reports some <concept_1> in the past, and had last <concept_2> on DATE_1.

Medical Concepts: [<concept_1>, <concept_2>]
---Example 4---

Here is the example to extract medical concepts from:

{section_heading}: {section_value}

Medical Concepts:
\end{lstlisting}

\begin{lstlisting}[language=promptlanguage,float=*,caption=Prompt for verifying concepts in metric computation.,label={prompt:verify_concepts}]
Given a snippet (snippet) from a medical dialogue summary and a corresponding list (list_a) of medical concepts extracted from that snippet, evaluate what medical concepts from a separate list (list_b) can be found in either list_a or snippet.

Note that on some occasions a medical concept from list_b may not be found in list_a, but can be appropriate to be present given the snippet. This could include rephrasings of medical concepts that are clinically equivalent (Ex: COVID and COVID-19).

---Example---
snippet: <snippet>
list_a: [<concept_1>, <concept_2>, <concept_3>, <concept_4>, <concept_5>, <concept_7>]
list_b: [<concept_0>, <concept_1>, <concept_3>, <concept_4>, <concept_5>, <concept_6>]

found_b: [<concept_1>, <concept_3>, <concept_4>, <concept_5>]
not_found_b: [<concept_0>, <concept_6>]

---Example---

Here is the snippet, list_a. Evaluate the medical concepts in list_b as above.

snippet: {snippet}
list_a: {list_a}
list_b: {list_b}

found_b:
\end{lstlisting}

\end{document}